\documentclass[a4,10pt]{article}
\usepackage{amsmath,amssymb, bm, statex, braket}
\usepackage{enumerate, fancybox}

\usepackage[dvipdfmx]{graphicx}

\allowdisplaybreaks[4]
\newcommand{\nn}{\nonumber \\}

\newcommand{\mcal}[1]{ \mathcal{#1} }
\newcommand{\mrm}[1]{ \mathrm{#1} }

\newcommand{\ave}[1]{{\langle #1 \rangle}}

\usepackage{amsthm, framed}
\newtheorem{proposition}{Proposition}
\newtheorem*{proposition*}{Proposition}
\newtheorem{corollary}{Corollary}
\newtheorem*{corollary*}{Corollary}

\begin{document}

\begin{center}
{\huge{\textsc{Mean-Field Inference\\ \vspace{2mm}in Gaussian Restricted Boltzmann Machine}}}\\

\ \\
\ \\
{\Large{Chako Takahashi and Muneki Yasuda\footnote{Corresponding author: muneki@yz.yamagata-u.ac.jp}}}\\
\ \\
Graduate School of Science and Engineering, Yamagata University, Japan
\end{center}

\subsubsection*{abstract:}
A Gaussian restricted Boltzmann machine (GRBM) is a Boltzmann machine
defined on a bipartite graph
and is an extension of usual restricted Boltzmann machines.
A GRBM consists of two different layers: a visible layer composed of
continuous visible variables and a hidden layer composed of discrete
hidden variables.
In this paper, we derive two different inference algorithms
for GRBMs based on the naive mean-field approximation (NMFA).
One is an inference algorithm for whole variables in a GRBM,
and the other is an inference algorithm for partial variables in a GBRBM.
We compare the two methods analytically and numerically
and show that the latter method is better.

\section{Introduction}

A restricted Boltzmann machine (RBM) is a statistical machine learning model that is defined on a bipartite graph~\cite{RBM1986,CD2002} 
and forms a fundamental component of deep learning~\cite{Hinton2006,RBM-Sci2006}. 
The increasing use of deep learning techniques in various fields 
is leading to a growing demand for the analysis of computational algorithms for RBMs. 
The computational procedure for RBMs is divided into two main stages: the learning stage and the inference stage. 
We train an RBM using an observed data set in the learning stage, 
and we compute some statistical quantities, e.g., expectations of variables, for the trained RBM in the inference stage. 
For the learning stage, many efficient algorithms, e.g., contrastive divergence~\cite{CD2002}, have been developed. 
On the other hand, algorithms for the inference stage have not witnessed much sophistication. 
Methods based on Gibbs sampling and the na\"ive mean-field approximation (NMFA) are mainly used for the inference stage, for example in references~\cite{RBMcFilter2007,TBM2013}. 
However, some new algorithms~\cite{BPinRBM2015a,BPinRBM2015b,TAPinRBM2015} based on advanced mean-field methods~\cite{Opper&Saad2001,Mezard&Montanari2009} have emerged in recent years. 

In this paper, we focus on a model referred to as the Gaussian restricted Boltzmann machine (GRBM) 
which is a slightly extended version of a Gaussian-Bernoulli restricted Boltzmann machine (GBRBM)~\cite{RBM-Sci2006,GBRBM_NIPS2007,GBRBM2011}. 
A GBRBM enables us to treat continuous data and is a fundamental component of a Gaussian-Bernoulli deep Boltzmann machine~\cite{DeepGBRBM2013}. 
In GBRBMs hidden variables are binary, whereas in GRBMs, hidden variables can take arbitrary discrete values.  
A statistical mechanical analysis for GBRBMs was presented in reference~\cite{HF&BM2012}.
For GRBMs, we study inference algorithms based on the NMFA. 
Since the NMFA is one of the most important foundations of advanced mean-field methods, 
gaining a deeper understanding of the NMFA for RBMs will provide us with some important insights into subsequent inference algorithms based on the advanced mean-field methods. 
For GRBMs, it is possible to obtain two different types of NMFAs: the NMFA for the whole system and 
the NMFA for a marginalized system. 
First, we derive the two approximations and then compare them analytically and numerically.
Finally, we show that the latter approximation is better.

The remainder of this paper is organized as follows.
The definition of GRBMs is presented in Section \ref{sec:GRBM}. 
The two different types of NMFAs are formulated in Sections \ref{sec:TypeI-MFA} and \ref{sec:TypeII-MFA}. 
Then, we compare the two methods analytically in Section \ref{sec:QualitativeComparison} and numerically in Section \ref{sec:QuantitativeComparison}, 
and we show that the NMFA for a marginalized system is better.
Finally, Section \ref{sec:conclud} concludes the paper.

\section{Gaussian Restricted Boltzmann Machine}
\label{sec:GRBM}

\begin{figure}[htb]
\begin{center}
\includegraphics[height=2.5cm]{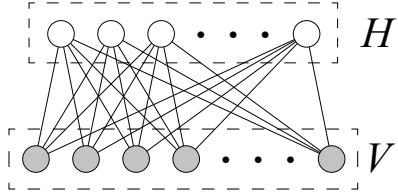}
\end{center}
\caption{Bipartite graph consisting of two layers: the visible layer $V$ and the hidden layer $H$.}
\label{fig:GRBM}
\end{figure}

Let us consider a bipartite graph consisting of two different layers: a visible layer and a hidden layer.
The continuous visible variables, $\bm{v} = \{v_i \in (-\infty, \infty) \mid i \in V\}$, are assigned to the vertices in the visible layer 
and the discrete hidden variables with a sample space $\mcal{X}$, $\bm{h} = \{ 
h_j \in 
\mcal{X} \mid j \in H\}$, are assigned to the vertices in the hidden layer, 
where $V$ and $H$ are the sets of vertices in the visible and the hidden layers, respectively.
Figure \ref{fig:GRBM} shows the bipartite graph. 
On the graph, we define the energy function as
\begin{align}
E(\bm{v},\bm{h} ; \theta):=\frac{1}{2}\sum_{i \in V}\frac{(v_i - b_i)^2}{\sigma_i^2}-\sum_{i \in V}\sum_{j \in H}\frac{w_{ij}}{\sigma_i^2}v_i h_j
-\sum_{j \in H}c_j h_j,
\label{eqn:energy-function}
\end{align}
where $b_i$, $\sigma_i^2$, $c_j$, and $w_{ij}$ are the parameters of the energy function and they are collectively denoted by $\theta$. 
Specifically, $b_i$ and $c_j$ are the biases for the visible and the hidden variables, respectively, $w_{ij}$ are the couplings between the visible and the hidden variables, and $\sigma_i^2$ are the parameters related to the variances of the visible variables.
The GRBM is defined by
\begin{align}
P(\bm{v}, \bm{h} \mid \theta):= \frac{1}{Z(\theta)}\exp\big( - E(\bm{v},\bm{h} ; \theta)\big)
\label{eqn:GRBM}
\end{align}
in terms of the energy function in Equation (\ref{eqn:energy-function}). 
Here, $Z(\theta)$ is the partition function defined by
\begin{align*}
Z(\theta):=\int \sum_{\bm{h}}\exp\big( - E(\bm{v},\bm{h} ; \theta)\big) d\bm{v}, 
\end{align*}
where $\int(\cdots)d\bm{v} = \int_{-\infty}^{\infty}\int_{-\infty}^{\infty}\cdots \int_{-\infty}^{\infty}(\cdots)dv_1 dv_2 \cdots dv_{|V|}$ is the multiple integration over all the possible realizations of the visible variables 
and $ \sum_{\bm{h}} = \sum_{h_1 \in \mcal{X}}\sum_{h_2 \in \mcal{X}}\cdots \sum_{h_{|H|} \in \mcal{X}}$ is the multiple summation over those of the hidden variables.　
When $\mcal{X} = \{+1,-1\}$, the GRBM corresponds to a GBRBM~\cite{GBRBM2011}.

The distribution of the visible variables conditioned with the hidden variables is
\begin{align}
P(\bm{v} \mid \bm{h}, \theta) = \prod_{i \in V} \mcal{N}(v_i \mid \mu_i(\bm{h}), \sigma_i^2),
\label{eqn:P_V|H}
\end{align}
where $\mcal{N}(x \mid \mu, \sigma^2)$ is the Gaussian over $x \in (- \infty, \infty)$ with mean $\mu$ and variance $\sigma^2$
and 
\begin{align}
\mu_i(\bm{h}) := b_i + \sum_{j \in H}w_{ij} h_j.
\label{eqn:def-mu_i}
\end{align}
On the other hand, the distribution of the hidden variables conditioned with the visible variables is
\begin{align}
P(\bm{h} \mid \bm{v}, \theta) =\prod_{j \in H}\frac{\exp(\lambda_j(\bm{v}) h_j)}{\sum_{h \in \mcal{X}} \exp(\lambda_j(\bm{v}) h)},
\label{eqn:P_H|V}
\end{align}
where
\begin{align}
\lambda_j(\bm{v}):=c_j + \sum_{i \in V}\frac{w_{ij}}{\sigma_i^2}v_i.
\label{eqn:def-lambda_j}
\end{align}
From Equations (\ref{eqn:P_V|H}) and (\ref{eqn:P_H|V}), it is ensured that if one layer is conditioned, 
the variables in the other are statistically independent of each other. This property is referred to as \textit{conditional independence}. 

The marginal distribution of the hidden variables is 
\begin{align}
P(\bm{h} \mid \theta)=\int P(\bm{v}, \bm{h} \mid \theta) d\bm{v}=\frac{z_H(\theta)}{Z(\theta)}\exp\Big( \sum_{j \in H}B_j h_j +\sum_{j \in H}D_j h_j^2+ \sum_{j < k \in H}J_{jk}h_jh_k\Big), 
\label{eqn:marginal-hidden}
\end{align}
where 
\begin{align*}
B_j :=c_j + \sum_{i \in V}\frac{b_i}{\sigma_i^2}w_{ij},\quad D_j := \frac{1}{2}\sum_{i \in V}\frac{w_{ij}^2}{\sigma_i^2}, \quad
J_{jk} := \sum_{i \in V} \frac{w_{ij}w_{ik}}{\sigma_i^2},
\end{align*}
and 
\begin{align*}
z_H(\theta):=\exp \Big( \frac{1}{2}\sum_{i \in V}\ln (2\pi \sigma_i^2)\Big).
\end{align*}
The sum $\sum_{j < k \in H}$ is the summation over all distinct pairs of hidden variables. 
The marginal distribution in Equation (\ref{eqn:marginal-hidden}) is the standard Boltzmann machine (or the multi-valued Ising model with anisotropic parameters) consisting of the hidden variables. 

Using Equations (\ref{eqn:P_V|H}) and (\ref{eqn:marginal-hidden}), the expectation of $v_i$ is expressed as
\begin{align}
\ave{v_i}&:=\int \sum_{\bm{h}} v_i P(\bm{v}, \bm{h} \mid \theta) d\bm{v}\nn
&\>= \sum_{\bm{h}}  \Big(\int v_i P(\bm{v} \mid \bm{h}, \theta)d\bm{v}\Big)P(\bm{h} \mid \theta)= b_i + \sum_{j \in H}w_{ij}\ave{h_j},
\end{align}
where $\ave{h_j}$ is the expectation of $h_j$. 
Therefore, it is found that the expectations of the visible variables are expressed in terms of the linear combination of the expectations of the hidden variables.

\section{Mean-Field Approximations for GRBMs}
\label{sec:MFAforGRBM}

In this section, for the GRBM defined in the previous section, we derive two different types of mean-field approximations: type I and type II mean-field approximations. 
The type I mean-field approximation is the NMFA for whole variables in the GRBM 
and the type II mean-field approximation is the NMFA for the marginal distribution in Equation (\ref{eqn:marginal-hidden}).
The strategy of the type I method is analogous with that in references~\cite{BPinRBM2015b,TAPinRBM2015}, 
and the strategy of the type II method is analogous with that in reference~\cite{BPinRBM2015a}.

\subsection{Type I Mean-Field Approximation for GRBMs}
\label{sec:TypeI-MFA}

Let us prepare a test distribution in the form 
\begin{align}
T_{1}(\bm{v}, \bm{h}):= \Big(\prod_{i \in V}q_i(v_i)\Big)\Big( \prod_{j\in H}u_j(h_j)\Big)
\label{eqn:test-distribution-I}
\end{align}
and define the Kullback-Leibler divergence (KLD) between the GRBM in Equation (\ref{eqn:GRBM}) and the test distribution as
\begin{align}
\mcal{K}_{1}[\{q_i, u_j\}]:=\int \sum_{\bm{h}}T_{1}(\bm{v}, \bm{h}) \ln \frac{T_{1}(\bm{v}, \bm{h})}{P(\bm{v}, \bm{h} \mid \theta)} d\bm{v}.
\label{eqn:KLD-I}
\end{align} 
The type I mean-field approximation is obtained by minimizing the KLD with respect to the test distribution. 
The KLD can be rewritten as
\begin{align}
\mcal{K}_{1}[\{q_i, u_j\}] = \mcal{F}_{1}[\{q_i, u_j\}] + \ln Z(\theta),
\label{eqn:KLD-I-trans}
\end{align}
where
\begin{align*}
\mcal{F}_{1}[\{q_i, u_j\}]:=\int \sum_{\bm{h}} E(\bm{v},\bm{h} ; \theta)T_{1}(\bm{v}, \bm{h})d\bm{v} 
+\int \sum_{\bm{h}} T_{\mrm{I}}(\bm{v}, \bm{h}) \ln T_{1}(\bm{v}, \bm{h})d\bm{v} 
\end{align*}
is the variational mean-field free energy of this approximation and it can be rewritten as
\begin{align}
&\mcal{F}_{1}[\{q_i, u_j\}]=\sum_{i\in V}\int_{-\infty}^{\infty}\frac{(v_i - b_i)^2}{2\sigma_i^2}q_i(v_i) d v_i
- \sum_{i\in V}\sum_{j\in H}\frac{w_{ij}}{\sigma_i^2}\int_{-\infty}^{\infty}v_iq_i(v_i) d v_i\sum_{h_j\in \mcal{X}}h_j u_j (h_j) \nonumber\\
&  - \sum_{j\in H}c_j\sum_{h_j \in \mcal{X}}h_j u_j (h_j)
+\sum_{i\in V}\int_{-\infty}^{\infty}q_i(v_i)\ln q_i(v_i) d v_i + \sum_{j\in H}\sum_{h_j \in \mcal{X}}u_j(h_j)\ln u_j(h_j).
\label{eqn:VariationalFreeEnergy-I}
\end{align}
Because $\ln Z(\theta)$ is constant with respect to the test distribution, we minimize the variational free energy instead of the KLD. 
By variational minimization of the variational free energy with respect to $q_i(v_i)$ and $u_j(h_j)$ under the normalizing constraints 
$\int_{-\infty}^{\infty} q_i(v_i) dv_i= 1$ and $\sum_{h_j \in \mcal{X}}u_j(h_j) = 1$, 
we obtain the resulting distributions as
\begin{align}
q_i^*(v_i) &= \mcal{N}(v_i \mid \mu_i(\bm{m}), \sigma_i^2),
\label{eqn:MF_qi-I}\\
u_j^*(h_j) &= \frac{\exp(\lambda_j(\bm{\nu}) h_j)}{\sum_{h \in \mcal{X}} \exp(\lambda_j(\bm{\nu}) h)},
\label{eqn:MF_uj-I}
\end{align}
where $\bm{\nu} = \{\nu_i \mid i \in V\}$ and $\bm{m} = \{m_j \mid j \in H\}$ are the expectations defined by
\begin{align}
\nu_i &:= \int_{-\infty}^{\infty} v_i q_i^*(v_i) dv_i= \mu_i(\bm{m}),
\label{eqn:def-nu_i-I}\\
m_j &:= \sum_{h_j \in \mcal{X}}h_j u_j^*(h_j).
\label{eqn:def-m_j-I}
\end{align}
The functions $\mu_i$ and $\lambda_j$ are respectively defined in Equations (\ref{eqn:def-mu_i}) and (\ref{eqn:def-lambda_j}). 
In the mean-field approximation, the distributions in Equations (\ref{eqn:MF_qi-I}) and (\ref{eqn:MF_uj-I}) are regarded as the mean-field approximation of the GRBM: 
$P(\bm{v}, \bm{h} \mid \theta) \approx T_{1}^*(\bm{v}, \bm{h}) = (\prod_{i \in V}q_i^*(v_i))(\prod_{j\in H}u_j^*(h_j))$. 
Therefore, $\bm{\nu}$ and $\bm{m}$, satisfying Equations (\ref{eqn:MF_uj-I})--(\ref{eqn:def-m_j-I}), are the approximate expectations of the visible and the hidden variables, respectively, because 
\begin{align*}
\ave{v_i}&=\int \sum_{\bm{h}} v_i P(\bm{v}, \bm{h} \mid \theta) d\bm{v} 
\approx  \int \sum_{\bm{h}} v_i T_{1}^*(\bm{v}, \bm{h}) d\bm{v} = \nu_i,\\
\ave{h_j}&=\int \sum_{\bm{h}} h_j P(\bm{v}, \bm{h} \mid \theta) d\bm{v} 
\approx  \int \sum_{\bm{h}} h_j T_{1}^*(\bm{v}, \bm{h}) d\bm{v} = m_j.
\end{align*}
By numerically solving the mean-field equations in Equations (\ref{eqn:MF_uj-I})--(\ref{eqn:def-m_j-I}) using a method of successive substitution for example, 
we can obtain the values of $\bm{\nu}$ and $\bm{m}$.

\subsection{Type II Mean-Field Approximation for GRBMs}
\label{sec:TypeII-MFA}

In the type II mean-field approximation, we use a test distribution in the form of 
\begin{align}
T_{2}(\bm{v}, \bm{h}):= P(\bm{v}\mid  \bm{h},\theta)\prod_{j\in H}u_j(h_j),
\label{eqn:test-distribution-II}
\end{align}
where $P(\bm{v}\mid  \bm{h},\theta)$ is the conditional distribution in Equation (\ref{eqn:P_V|H}). 
The KLD between the test distribution and the GRBM is expressed as
\begin{align}
\mcal{K}_{2}[\{u_j\}]:=\int \sum_{\bm{h}}T_{2}(\bm{v}, \bm{h}) \ln \frac{T_{2}(\bm{v}, \bm{h})}{P(\bm{v}, \bm{h} \mid \theta)} d\bm{v}
=\sum_{\bm{h}}\Big(\prod_{j\in H}u_j(h_j)\Big) \ln \frac{\prod_{j\in H}u_j(h_j)}{P(\bm{h} \mid \theta)},
\label{eqn:KLD-II}
\end{align} 
where $P(\bm{h} \mid \theta)$ is the marginal distribution in Equation (\ref{eqn:marginal-hidden}). 
The KLD can be rewritten as 
\begin{align}
\mcal{K}_{2}[\{u_j\}] = \mcal{F}_{2}[\{u_j\}] + \ln Z(\theta),
\label{eqn:KLD-II-trans}
\end{align}
where
\begin{align*}
\mcal{F}_{2}[\{u_j\}]:= \sum_{\bm{h}} E_H(\bm{h} ; \theta)\prod_{j\in H}u_j(h_j) +\sum_{\bm{h}}\Big(\prod_{j\in H}u_j(h_j)\Big) \ln \prod_{j\in H}u_j(h_j) - \ln z_H(\theta)
\end{align*}
is the variational mean-field free energy of this approximation and 
\begin{align*}
E_H(\bm{h} ; \theta):=-\sum_{j \in H}B_j h_j -\sum_{j \in H}D_j h_j^2- \sum_{j < k \in H}J_{jk}h_jh_k
\end{align*}
is the energy function of the marginal distribution in Equation (\ref{eqn:marginal-hidden}). 
This variational mean-field free energy can be rewritten as
\begin{align}
&\mcal{F}_{2}[\{u_j\}]= -\sum_{j \in H}B_j \sum_{h_j \in \mcal{X}} h_j u_j(h_j) -\sum_{j \in H}D_j \sum_{h_j \in \mcal{X}} h_j^2 u_j(h_j)\nn
&- \sum_{j < k \in H}J_{jk}\sum_{h_j \in \mcal{X}} h_j u_j(h_j)\sum_{h_k \in \mcal{X}} h_k u_j(h_k) +\sum_{j \in H}\sum_{h_j \in \mcal{X}}u_j(h_j) \ln u_j(h_j) - \ln z_H(\theta).
\label{eqn:VariationalFreeEnergy-II}
\end{align}

By variational minimization of the variational free energy in Equation (\ref{eqn:VariationalFreeEnergy-II}) under the normalizing constraints 
$\sum_{h_j \in \mcal{X}}u_j(h_j) = 1$, we obtain 
\begin{align}
u_j^{\dagger}(h_j) = \frac{\exp \big(B_j h_j + D_j h_j^2 + \sum_{k \in H 
\setminus\{j\}}J_{jk}m_k^{\dagger} h_j \big)}
{\sum_{h \in \mcal{X}} \exp \big(B_j h + D_j h^2 + \sum_{k \in H \setminus\{j\}}J_{jk}m_k^{\dagger} h \big)},
\label{eqn:MF_uj-II}
\end{align}
where $\bm{m}^{\dagger} = \{m_j^{\dagger} \mid j \in H\}$ are the expectations defined by
\begin{align}
m_j^{\dagger} := \sum_{h_j \in \mcal{X}}h_j u_j^{\dagger}(h_j).
\label{eqn:def-m_j-II}
\end{align}
The resulting test distribution, $T_2^{\dagger}(\bm{v}, \bm{h}) = P(\bm{v}\mid  
\bm{h},\theta) \prod_{j\in H}u_j^{\dagger}(h_j)$, is regarded as the mean-field 
approximation of the GRBM in this approximation. 
Therefore, $\bm{m}^{\dagger}$, which satisfy Equations (\ref{eqn:MF_uj-II}) and 
(\ref{eqn:def-m_j-II}), are the approximate expectations of the hidden 
variables, because
\begin{align*}
\ave{h_j}=\int \sum_{\bm{h}} h_j P(\bm{v}, \bm{h} \mid \theta) d\bm{v} 
\approx  \int \sum_{\bm{h}} h_j T_{2}^{\dagger}(\bm{v}, \bm{h}) d\bm{v} = 
m_j^{\dagger}.
\end{align*}
By solving the mean-field equations in Equations (\ref{eqn:MF_uj-II}) and (\ref{eqn:def-m_j-II}), we can obtain the values of $\bm{m}^{\dagger}$.
The approximate expectations of the visible variables, $\bm{\nu}^{\dagger} = \{\nu_i^{\dagger} \mid i \in V\}$, can be obtained in terms of $\bm{m}^{\dagger}$ as
\begin{align}
\ave{v_i}\approx \nu_i^{\dagger} 
:=  \int \sum_{\bm{h}} v_i T_{2}^{\dagger}(\bm{v}, \bm{h}) d\bm{v} = \mu_i(\bm{m}^{\dagger}).
\label{eqn:ApproxExpect_Visible-II}
\end{align}
Using Equations (\ref{eqn:MF_uj-II})--(\ref{eqn:ApproxExpect_Visible-II}) we can rewrite the above mean-field equations as
\begin{align}
\nu_i^{\dagger} &= \mu_i(\bm{m}^{\dagger}),
\label{eqn:MFE-nu_i-II}\\
m_j^{\dagger} &= \sum_{h_j \in \mcal{X}}h_j y_j(h_j),
\label{eqn:MFE-m_j-II}
\end{align}
where
\begin{align*}
y_j(h_j):=\frac{\exp \big\{\lambda_j(\bm{\nu}^{\dagger}) h_j  -\sum_{i \in V}(w_{ij} / \sigma_i)^2\big(m_j^{\dagger} - h_j/2\big) h_j\big\}}
{\sum_{h \in \mcal{X}} \exp \big\{\lambda_j(\bm{\nu}^{\dagger}) h  -\sum_{i \in V}(w_{ij} / \sigma_i)^2\big(m_j^{\dagger} - h/2\big) h\big\}}.
\end{align*}
The values of $\bm{\nu}^{\dagger}$ and $\bm{m}^{\dagger}$ can also be obtained by numerically solving the mean-field equations in Equations (\ref{eqn:MFE-nu_i-II}) and (\ref{eqn:MFE-m_j-II}) instead of solving those in Equations (\ref{eqn:MF_uj-II})--(\ref{eqn:ApproxExpect_Visible-II}). 
The order of the computational cost of solving the mean-field equations is the same as that of the type I mean-field approximation presented in Section \ref{sec:TypeI-MFA}.

\section{Comparison of Two Mean-Field Methods}
\label{sec:ComparisonTwoMethods}

In Sections \ref{sec:TypeI-MFA} and \ref{sec:TypeII-MFA}, we derived two different mean-field approximations for the GRBM: the type I and the type II mean-field approximations. 
Both the approximations are constructed on the basis of the NMFA. 
Now, we are interested in which approximation is better.
Intuitively, the type II mean-field approximation seems to be better, 
because the number of variables that imposes the mean-field assumption, namely, the factorizable assumption of the distribution, in the type II mean-field approximation is less than that in the type I mean-field approximation. 
In this section, we qualitatively and quantitatively compare the two approximations and show that our intuitive prediction is valid.

\subsection{Qualitative Comparison}
\label{sec:QualitativeComparison}

Before discussing the qualitative relationship between the mean-field approximations, 
we provide a general theorem that will be an important basis of our final results in this section. 
Let us consider continuous or discrete random variables $\bm{x} = \{x_i \mid i = 1,2,\ldots,n\}$ 
and divide the variables into two different sets: $\bm{x} = \bm{x}_A \cup \bm{x}_B$. 
We define a distribution $P(\bm{x})$ over the random variables, and define two kinds of test distribution for the distribution as
$T_{\mrm{all}}(\bm{x}):=Q(\bm{x}_A)U(\bm{x}_B)$ and $T_{\mrm{part}}(\bm{x}):=P(\bm{x}_A \mid \bm{x}_B)U(\bm{x}_B)$.
$P(\bm{x}_A \mid \bm{x}_B)$ is the conditional distribution of $P(\bm{x})$, and $Q(\bm{x}_A)$ and $U(\bm{x}_B)$ are distributions over $\bm{x}_A$ and $\bm{x}_B$, respectively. 
For the test distributions $T_{\mrm{all}}(\bm{x})$ and $T_{\mrm{part}}(\bm{x})$, we define the KLDs as
\begin{align}
\mcal{K}_{\mrm{all}}[Q,U] :=\sum_{\bm{x}}T_{\mrm{all}}(\bm{x}) \ln \frac{T_{\mrm{all}}(\bm{x})}{P(\bm{x})}
\label{eqn:KLD-all}
\end{align} 
and
\begin{align}
\mcal{K}_{\mrm{part}}[U] :=\sum_{\bm{x}}T_{\mrm{part}}(\bm{x}) \ln \frac{T_{\mrm{part}}(\bm{x})}{P(\bm{x})},
\label{eqn:KLD-part}
\end{align} 
respectively, where the sum $\sum_{\bm{x}} = \sum_{x_1}\sum_{x_2}\cdots \sum_{x_n}$ is the multiple summation over all the possible realizations of $\bm{x}$. 
If some variables are continuous, the corresponding summations become integrations.
Under the setting, we obtain the following proposition.
\begin{leftbar}
\begin{proposition} 
For any distribution $P(\bm{x})$ over $n$ random variables $\bm{x} = \{x_i \mid 
i = 1,2,\ldots,n\}$ with any sample spaces, the inequality
\begin{align*}
\min_{Q, U}\mcal{K}_{\mrm{all}}[Q,U]\geq \min_U \mcal{K}_{\mrm{part}}[U]
\end{align*}
is ensured, where $\mcal{K}_{\mrm{all}}[Q,U]$ and $\mcal{K}_{\mrm{part}}[U]$ are the KLDs defined in Equations (\ref{eqn:KLD-all}) and (\ref{eqn:KLD-part}), respectively. 
\label{prop:KLD-general}
\end{proposition}
\end{leftbar}
\begin{proof}
From Equations (\ref{eqn:KLD-all}) and (\ref{eqn:KLD-part}), $\mcal{K}_{\mrm{all}}[Q,U]$ can be rewritten as
\begin{align*}
\mcal{K}_{\mrm{all}}[Q,U] = \sum_{\bm{x}}Q(\bm{x}_A)U(\bm{x}_B) \ln \frac{Q(\bm{x}_A)}{P(\bm{x}_A \mid \bm{x}_B)} + \mcal{K}_{\mrm{part}}[U].
\end{align*}
By using this expression, the inequality
\begin{align}
\min_{Q,U}\mcal{K}_{\mrm{all}}[Q,U]&=\sum_{\bm{x}}Q^*(\bm{x}_A)U^*(\bm{x}_B) \ln \frac{Q^*(\bm{x}_A)}{P(\bm{x}_A \mid \bm{x}_B)} + \mcal{K}_{\mrm{part}}[U^*]\nn
&\geq \sum_{\bm{x}}Q^*(\bm{x}_A)U^*(\bm{x}_B) \ln \frac{Q^*(\bm{x}_A)}{P(\bm{x}_A \mid \bm{x}_B)} + \min_U \mcal{K}_{\mrm{part}}[U]
\label{eqn:inequality-1-prop1}
\end{align}
is obtained, where $Q^*(\bm{x}_A)$ and $U^*(\bm{x}_B)$ are the distributions that minimize $\mcal{K}_{\mrm{all}}[Q,U]$.
By using the inequality $\ln X \leq X - 1$ for $X \geq  0$, we obtain
\begin{align}
- \sum_{\bm{x}}Q^*(\bm{x}_A)U^*(\bm{x}_B) \ln \frac{P(\bm{x}_A \mid \bm{x}_B)}{Q^*(\bm{x}_A)}
\geq \sum_{\bm{x}}Q^*(\bm{x}_A)U^*(\bm{x}_B) \Big(1 - \frac{P(\bm{x}_A \mid \bm{x}_B)}{Q^*(\bm{x}_A)}\Big) = 0.
\label{eqn:inequality-2-prop1}
\end{align}
From Equations (\ref{eqn:inequality-1-prop1}) and (\ref{eqn:inequality-2-prop1}), the proposition is obtained.
\end{proof}
 
In Proposition \ref{prop:KLD-general}, by regarding $\bm{x}_A$ and $\bm{x}_B$ as $\bm{v}$ and $\bm{h}$, respectively, and by regarding $P(\bm{x})$ as the GRBM,
we immediately obtain the following corollary.
\begin{leftbar}
\begin{corollary}
For the GRBM in Equation (\ref{eqn:GRBM}), the inequality
\begin{align*}
\min_{\{q_i, u_j\}}\mcal{K}_{1}[\{q_i, u_j\}] \geq  \min_{\{u_j\}} \mcal{K}_{2}[\{u_j\}]
\end{align*}
is ensured, where $\mcal{K}_{1}[\{q_i, u_j\}]$ and $\mcal{K}_{2}[\{u_j\}]$ are the KLDs defined in Equations (\ref{eqn:KLD-I}) and (\ref{eqn:KLD-II}).
\label{prop:KLD}
\end{corollary}
\end{leftbar}
A KLD is regarded as a measure of the distance between two different distributions. 
Corollary \ref{prop:KLD} suggests that the mean-field distribution obtained by the type II mean-field approximation is closer to the GRBM than that obtained by the type I mean-field approximation from the viewpoint of the KLD. 

We can obtain the following proposition for free energies. 
\begin{leftbar}
\begin{proposition}
For the GRBM in Equation (\ref{eqn:GRBM}), the inequality
\begin{align*}
F_1(\theta) \geq F_2(\theta) \geq  F(\theta)
\end{align*}
is ensured, where $F_1(\theta)$ and $F_2(\theta)$, defined by $F_1(\theta):= \min_{\{q_i, u_j\}}\mcal{F}_{1}[\{q_i, u_j\}]$ and $F_2(\theta):= \min_{\{u_j\}}\mcal{F}_{2}[\{u_j\}]$, are the mean-field free energies obtained by the type I and the type II mean-field approximations, respectively, and where $F(\theta) := - \ln Z(\theta)$ is the true free energy of the GRBM.
\label{prop:FreeEnergy}
\end{proposition}
\end{leftbar}
\begin{proof}
Since a KLD is nonnegative, from Equations (\ref{eqn:KLD-I-trans}) and (\ref{eqn:KLD-II-trans}), we obtain
\begin{align}
F_1(\theta) \geq F(\theta), \quad F_2(\theta) \geq F(\theta).
\label{eqn:inequality-1-prop3}
\end{align}
From Corollary \ref{prop:KLD} and Equations (\ref{eqn:KLD-I-trans}) and (\ref{eqn:KLD-II-trans}), we have
\begin{align}
F_1(\theta) \geq  F_2(\theta)
\label{eqn:inequality-2-prop3}
\end{align}
From Equations (\ref{eqn:inequality-1-prop3}) and (\ref{eqn:inequality-2-prop3}), we obtain the proposition.
\end{proof}
From this proposition, it is guaranteed that the mean-field free energy obtained by the type II mean-field approximation is closer to the true free energy than that obtained by the type I mean-field approximation.

\subsection{Quantitative Comparison}
\label{sec:QuantitativeComparison}

In this section, we quantitatively compare the two mean-field approximations through numerical experiments. 
In the numerical experiments, we use a GRBM with 24 visible variables and 12 hidden 
variables. 
Because the size of the used GRBM is small, we can evaluate the exact values of 
its free energy and expectations.
In the following experiments, we generate the values of the biases, $b_i$ and 
$c_j$, and 
of the couplings $w_{ij}$ from Gaussian distributions, 
and we fix the values of all $\sigma_i^2$ as one. 

\begin{figure}[htb]
\begin{center}
\includegraphics[height=5.0cm]{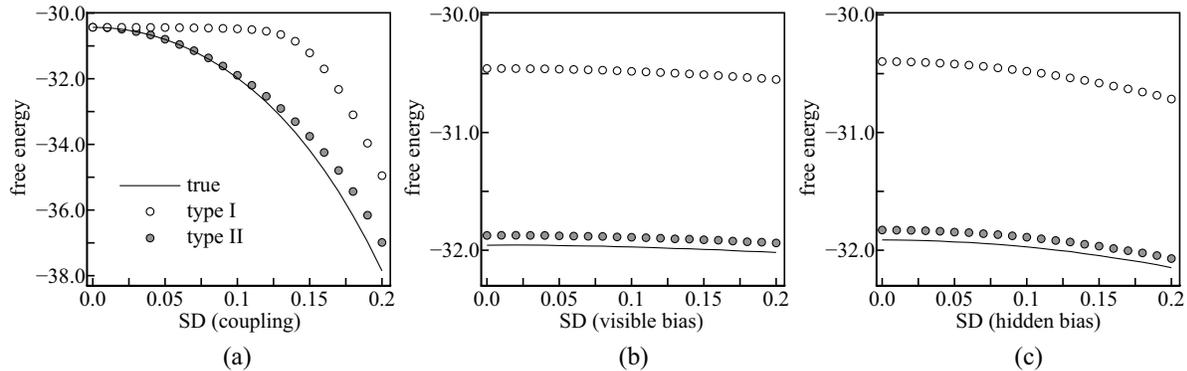}
\end{center}
\caption{Dependency of the free energies on the standard deviation of (a) the couplings $\bm{w}$, (b) the biases $\bm{b}$, and (c) the biases $\bm{c}$, when $\mcal{X} = \{-1,+1\}$.}
\label{fig:FE_2}
\end{figure}
\begin{figure}[htb]
\begin{center}
\includegraphics[height=5.0cm]{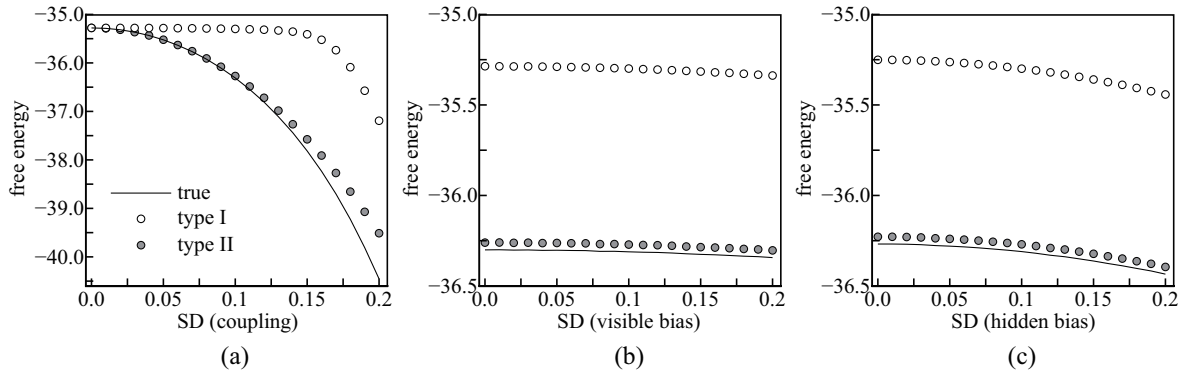}
\end{center}
\caption{Dependency of the free energies on the standard deviation of (a) the couplings $\bm{w}$, (b) the biases $\bm{b}$, and (c) the biases $\bm{c}$, when $\mcal{X} = \{-1,0,+1\}$.}
\label{fig:FE_3}
\end{figure}
Figures \ref{fig:FE_2} and \ref{fig:FE_3} show the dependencies of the three free energies, the type I mean-
field free energy $F_1(\theta)$, the type II mean-field free energy $F_2(\theta)
$, and the true free energy $F(\theta)$, on the parameters when $\mcal{X} = \{-1,+1\}$ and $\mcal{X} = \{-1,0,+1\}$, respectively. 
Since the partition function of the marginal distribution in Equation (\ref{eqn:marginal-hidden}) is $Z(\theta) / z_H(\theta)$, the true free energy can be evaluated by performing the following multiple summation.
\begin{align*}
F(\theta)= - \frac{1}{2}\sum_{i \in V}\ln (2\pi \sigma_i^2) - \ln \sum_{\bm{h}}\exp\Big( \sum_{j \in H}B_j h_j +\sum_{j \in H}D_j h_j^2+ \sum_{j < k \in H}J_{jk}h_jh_k\Big).
\end{align*}
The mean-field free energies, $F_1(\theta)$ and $F_2(\theta)$, are obtained by substituting the solutions to the mean-field equations of the type I and type II methods, i.e., $\{\nu_i, m_j\}$ and $\{\nu_i^{\dagger} , m_j^{\dagger}\}$, into Equations (\ref{eqn:VariationalFreeEnergy-I}) and (\ref{eqn:VariationalFreeEnergy-II}), respectively. 
Each plot in Figures \ref{fig:FE_2} and \ref{fig:FE_3} is the average over 10000 trials, and the parameters, $\bm{b}$, $\bm{c}$, and $\bm{w}$, used in the experiments were generated as follows.
For figures \ref{fig:FE_2}(a) and \ref{fig:FE_3}(a), they were independently drawn from 
$\mcal{N}(b_i \mid 0, 0.1^2)$, $\mcal{N}(c_j \mid 0, 0.1^2)$, and $\mcal{N}(w_{ij} \mid 0, \mrm{SD}^2)$, respectively.
For figures \ref{fig:FE_2}(b) and \ref{fig:FE_3}(b), they were independently drawn from 
$\mcal{N}(b_i \mid 0, \mrm{SD}^2)$, $\mcal{N}(c_j \mid 0, 0.1^2)$, and $\mcal{N}(w_{ij} \mid 0, 0.1^2)$, respectively.
For figures \ref{fig:FE_2}(c) and \ref{fig:FE_3}(c), they were independently drawn from 
$\mcal{N}(b_i \mid 0, 0.1^2)$, $\mcal{N}(c_j \mid 0, \mrm{SD}^2)$, and $\mcal{N}(w_{ij} \mid 0, 0.1^2)$, respectively.
One can observe that the results shown in Figures \ref{fig:FE_2} and \ref{fig:FE_3} are consistent with the theoretical result presented in Proposition \ref{prop:FreeEnergy}.

\begin{figure}[htb]
\begin{center}
\includegraphics[height=5.0cm]{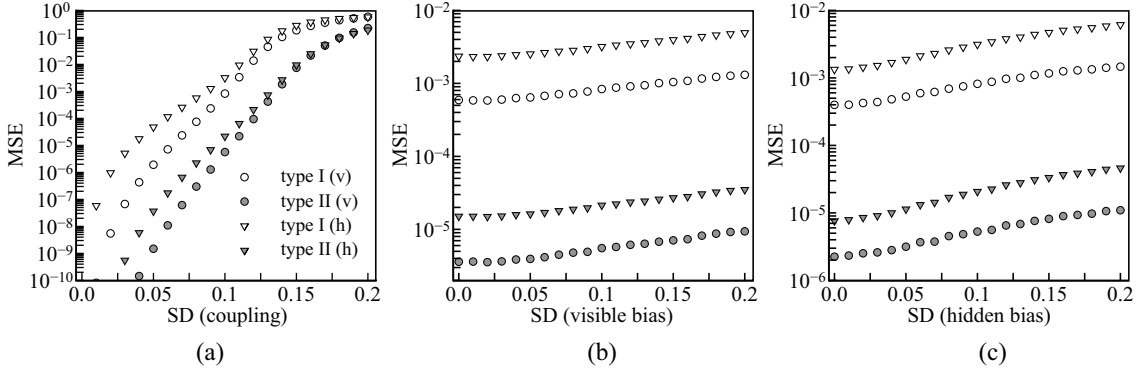}
\end{center}
\caption{Dependency of the MSEs of the expectations on the standard deviation of (a) the couplings $\bm{w}$, (b) the biases $\bm{b}$, and (c) the biases $\bm{c}$, when $\mcal{X} = \{-1,+1\}$.}
\label{fig:MSE_2}
\end{figure}
\begin{figure}[htb]
\begin{center}
\includegraphics[height=5.0cm]{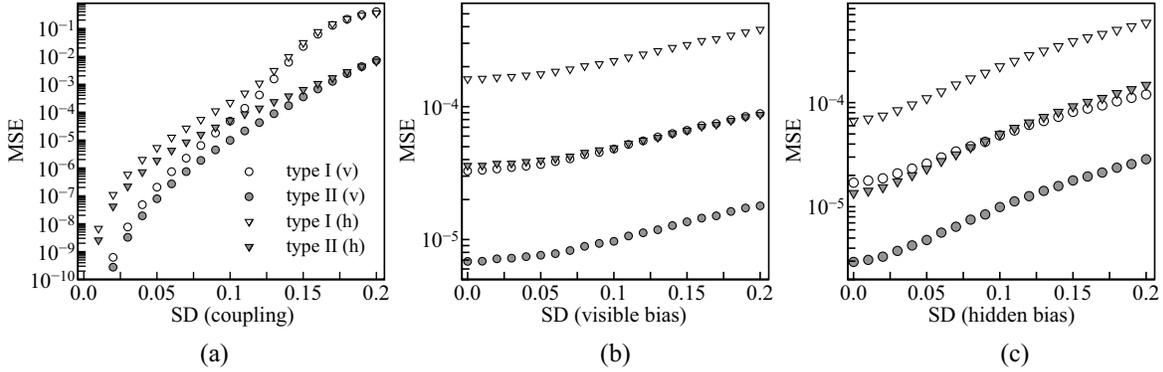}
\end{center}
\caption{Dependency of the MSEs of the expectations on the standard deviation of (a) the couplings $\bm{w}$, (b) the biases $\bm{b}$, and (c) the biases $\bm{c}$, when $\mcal{X} = \{-1,0,+1\}$.}
\label{fig:MSE_3}
\end{figure}
Figures \ref{fig:MSE_2} and \ref{fig:MSE_3} show the dependencies of the mean square errors (MSEs) between the exact expectations and the mean-field solutions on the parameters when $\mcal{X} = \{-1,+1\}$ and $\mcal{X} = \{-1,0,+1\}$, respectively. 
The plots ``type I (h)'' and ``type I (v)'' are the MSEs between $\ave{h_j}$ and $m_j$ and between $\ave{v_i}$ and $\nu_i$, respectively, that is, 
$|H|^{-1}\sum_{j \in H}(\ave{h_j} - m_j)^2$ and $|V|^{-1}\sum_{i \in V}(\ave{v_i} - \nu_i)^2$, respectively.
The plots ``type II (h)'' and ``type II (v)'' are the MSEs between $\ave{h_j}$ and $m_j^{\dagger}$ and between $\ave{v_i}$ and $\nu_i^{\dagger}$, respectively, that is, 
$|H|^{-1}\sum_{j \in H}(\ave{h_j} - m_j^{\dagger})^2$ and $|V|^{-1}\sum_{i \in V}(\ave{v_i} - \nu_i^{\dagger})^2$, respectively. 
Each plot in Figures \ref{fig:MSE_2} and \ref{fig:MSE_3} is the average over 10000 trials, 
and the parameters, $\bm{b}$, $\bm{c}$, and $\bm{w}$, used in the experiments were generated in the same manner as that for Figures \ref{fig:FE_2} and \ref{fig:FE_3}. 
We can observe that the type II method gives better approximations than the type I method.

\section{Conclusion}
\label{sec:conclud}

In this paper, we derived two different types of NMFAs, the type I and the type II methods, for GRBMs 
and compared them analytically and numerically. 
Further, we presented propositions and a corollary that guarantee that the type II method provides (1) a lower value of the KLD than the type I method 
and (2) mean-field free energy that is closer to the true free energy than that provided by the type I method. 
Moreover, in our numerical experiments, we observed that the expectations obtained by the type II method are more accurate than those obtained by the type I method. 
Since the orders of the computational costs of the two methods are the same, 
we can conclude that the type II method is better than the type I method, 
that is, we should apply the NMFA to a marginalized system rather than to the whole system.

Although the statements presented in this paper were made for only the NMFA,  
we expect that the insights obtained in this paper can be extended to advanced mean-field methods. 
From this perspective, the results obtained in this paper implicitly support the validity of the method presented in reference~\cite{BPinRBM2015a}. 
We are now interested in the application of more sophisticated mean-field methods, such as the adaptive TAP method~\cite{AdaTAP2001} 
and susceptibility propagation~\cite{Yasuda2013}, to GRBMs. 
In particular, we believe that the application of the adaptive TAP method is important, 
because as mentioned in reference~\cite{HF&BM2012} 
GRBMs are strongly related to a Hopfield-type of system and the adaptive TAP method can be justified in such a system.
This will be addressed in our future studies.

\subsection*{acknowledgment}
This work was partially supported by CREST, Japan Science and
Technology Agency and by JSPS KAKENHI Grant Numbers 15K00330, 25280089, and 15H03699.

\bibliographystyle{jpsj}
\bibliography{citations}

\end{document}